\documentclass{article}
\usepackage{iclr2027_conference,times}
\usepackage[utf8]{inputenc}
\usepackage[T1]{fontenc}
\usepackage{hyperref}
\usepackage{url}
\usepackage{xurl}
\usepackage{booktabs}
\usepackage{amsfonts}
\usepackage{amssymb}
\usepackage{amsmath}
\usepackage{nicefrac}
\usepackage{microtype}
\usepackage[table]{xcolor}
\definecolor{grouplavender}{RGB}{239,235,249}
\usepackage{graphicx}
\usepackage{xspace}
\usepackage{pifont}
\usepackage{fontawesome5}
\usepackage{multirow}
\usepackage{booktabs}
\usepackage{amssymb}
\usepackage{makecell}
\usepackage{wrapfig}
\usepackage{algorithm}
\usepackage{algpseudocode}
\newcommand{\methodname}{PlayWorld}
\newcommand{\method}{\methodname\xspace}

\newcommand{\player}{Agent Player\xspace}
\DeclareRobustCommand{\cameraicon}{\faIcon{video}}
\DeclareRobustCommand{\actionicon}{\faIcon{gamepad}}

\newcommand{\eg}{\textit{e.g.}\xspace}

\title{\centering\fontsize{16}{19}\selectfont
\method: Benchmarking World Models\\
with Agent Players over Long-Horizon Objectives}

\author{\normalfont
  Kaixin Ding\textsuperscript{$\rm 1^{\dagger}$} \quad
  Xi Chen\textsuperscript{$\rm 1^{\ddagger}$} \quad
  Minghong Cai\textsuperscript{\rm 3} \quad
  Zhiyuan Xu\textsuperscript{\rm 1} \quad
  Yiyang Wang\textsuperscript{\rm 1} \quad
  Yuxiang Lu\textsuperscript{\rm 1} \\
  Junyi Li\textsuperscript{\rm 1} \quad
  Shuyang Chen\textsuperscript{\rm 4} \quad
  Yuan Gao\textsuperscript{\rm 2} \quad
  Xin Tao\textsuperscript{\rm 2} \quad
  Pengfei Wan\textsuperscript{\rm 2} \quad
  Hengshuang Zhao\textsuperscript{\rm 1\faEnvelope} \\[5pt]
  \textsuperscript{\rm 1}The University of Hong Kong \quad
  \textsuperscript{\rm 2}Kling Team, Kuaishou Technology \\ 
  \textsuperscript{\rm 3}The Chinese University of Hong Kong \quad
  \textsuperscript{\rm 4}Zhejiang University \\[5pt]
  \href{https://kxding.github.io/project/PlayWorld/}{\faHome\ Project Page} \quad
  \href{https://huggingface.co/datasets/jocelynd/playworld-bench}{\faDatabase\ Dataset} \quad
  \href{https://github.com/kxding/PlayWorld}{\faGithub\ GitHub} \quad
  \href{https://huggingface.co/spaces/jocelynd/PlayWorld-Leaderboard}{\faTrophy\ Leaderboard}
}

\iclrfinalcopy

\begin{document}

\maketitle
\let\thefootnote\relax
\footnotetext{\noindent\hspace{-1em}$\dagger$ Work done at Kling Team, Kuaishou Technology.}
\footnotetext{\noindent\hspace{-1em}$\ddagger$ Project Lead. \quad \faEnvelope\ Corresponding author.}
\lhead{Preprint}

\begin{figure}[h]\centering
\vspace{-1em}
\includegraphics[width=\textwidth]{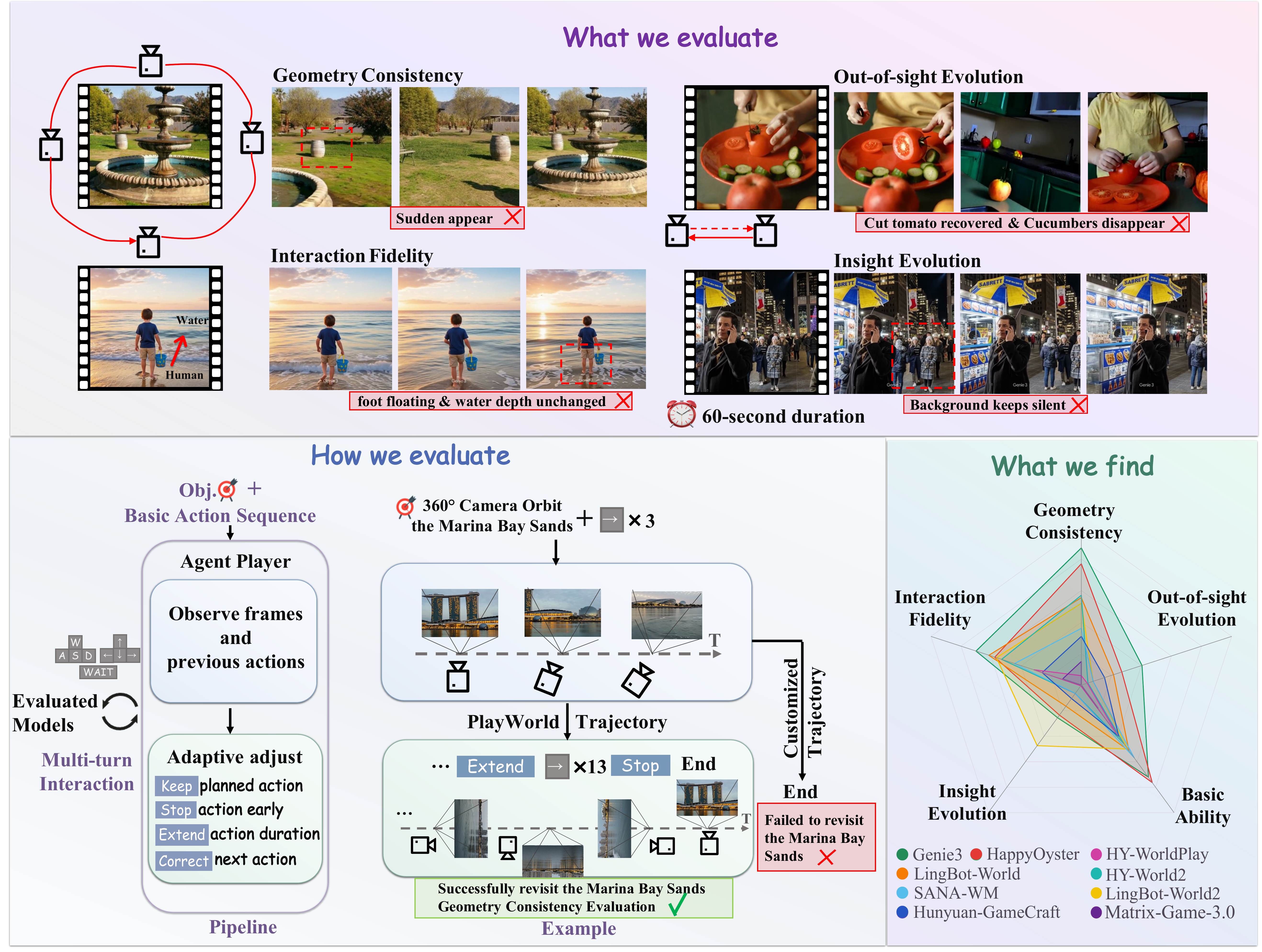}
\vspace{-1.8em}
\caption{
\textbf{\method evaluates world models from the perspective of a human player.}
We define long-horizon objectives such as ``turning around 360 degrees'' or ``walking into the water'', and use an Agent Player to interact with the world models.
We then evaluate them from multiple perspectives, including geometry consistency, interaction fidelity, and persistent state evolution.}
\label{fig:teaser}
\end{figure}

\begin{abstract}

Video world models simulate future states conditioned on current observations and user actions.
Recent systems have demonstrated impressive video consistency and action controllability over long sequences.
However, fairly comparing these interactive models remains challenging.
In practice, a human player typically evaluates a world model by pursuing long-horizon objectives through interaction.
For example, a user may turn around 360 degrees to see whether the environment remains consistent, or walk into the water and inspect whether realistic water ripples are generated.
The action sequence required to achieve the same objective may vary substantially between models, making fixed action-conditioned evaluation unsuitable for cross-model comparison.
To address this, we employ multi-modal Agent Players to interact with world models toward specified long-horizon objectives.
Building on this paradigm, we introduce \method, a benchmark providing 171 scenarios, each with a specified objective.
To evaluate performance thoroughly, we assess models along four core dimensions: geometry consistency, interaction fidelity, out-of-sight evolution, and insight evolution.
In addition, we incorporate basic ability metrics for video quality and controllability.
Experiments across nine state-of-the-art world models reveal that current models remain unreliable on long-horizon interactive objectives, particularly in maintaining spatial consistency and persistent state evolution.
Code and data are available at \url{https://kxding.github.io/project/PlayWorld/}

\end{abstract}

\section{Introduction}
\label{sec:introduction}

General video generation models have advanced rapidly in recent years, achieving remarkable visual quality and photorealism~\citep{_VideoGenerationModels_2024,wan,seedance_Seedance20Advancing_2026}.
Video world models~\citep{bruce_GenieGenerativeInteractive_2024,genie3,happyoyster} go a step further by simulating interactive environments that users can explore while maintaining causally consistent evolution across space and time.
This turns them into open worlds that users can freely interact with.

As a rapidly growing body of work pursues this goal~\citep{astra,li_HunyuanGameCraftHighdynamicInteractive_2025,he_Matrixgame20Opensource_2025,wang_MatrixGame30RealTime_2026,gao2026lingbotworld2,zhu2026sanawm,worldplay2025}, evaluation has become an increasingly important challenge.
General video-generation benchmarks such as VBench~\citep{vbench} assess aesthetics, temporal coherence, and text alignment, but do not measure whether a model responds correctly to user actions.
World-model benchmarks further evaluate 3D consistency, memory consistency, interaction, and out-of-sight state evolution~\citep{worldscore,ma_OutSightOut_2026,chen2026memobench,xu2026worldmark,wu_OmniWorldBenchComprehensiveInteractionCentric_2026}.
However, these benchmarks typically rely on customized trajectories predefined for individual evaluation cases, as summarized in Tab.~\ref{tab:criteria}.

This evaluation paradigm does not fully match how human players judge interactive world models.
In practice, a human user usually evaluates a world model by pursuing a high-level, long-horizon objective rather than by checking whether a predefined action trajectory is followed.
For example, a customized trajectory may specify three $\rightarrow$ commands with the intent of rotating the camera by $360^\circ$ and returning to the initial view of Marina Bay Sands, as illustrated in Fig.~\ref{fig:teaser}.
Because action granularity varies across models, the same three commands may complete a full rotation in one rollout but only a partial turn in another.
Marina Bay Sands may therefore leave and re-enter the view in some rollouts but fail to reappear in others, making the resulting geometry consistency scores incomparable.
Such customized trajectories also cannot reliably complete more complex action patterns, such as orbiting a sculpture while maintaining a fixed viewing direction~(Fig.~\ref{fig:suite_stats}), because the required action duration and stopping point differ across models.
Moreover, web-based closed-source world models such as Genie 3~\citep{genie3} and \mbox{HappyOyster~\citep{happyoyster}} are accessible only through a web interface and require direct human interaction, which is labor-intensive and prone to bias.
These challenges call for an automated evaluation protocol that can interact with world models like a human player.

To address this challenge, we introduce \method, which evaluates world models using long-horizon objectives.
\method uses \player, a multi-modal agent that simulates a human player.
Each evaluation case provides a basic action sequence, which serves as a shared initial reference across world models.
Given this reference, the Agent Player observes the generated frames and action history at each step and adaptively adjusts execution through \textit{Keep}, \textit{Stop}, \textit{Extend}, \textit{Correct}, or \textit{End} decisions.
These decisions adapt the number and duration of actions to each model's response, enabling fairer comparison under the same objectives.
\method further introduces a VQA rubric verifier for open-ended interactive scenarios, covering four core world-model capabilities: geometry consistency, interaction fidelity, out-of-sight evolution, and insight evolution.
The overall evaluation framework assesses model behavior over rollouts lasting approximately 10 to 60 seconds using continuous W/A/S/D, $\uparrow$/$\downarrow$/$\leftarrow$/$\rightarrow$, and WAIT controls.
The benchmark contains 171 human-annotated cases evaluated across nine representative world models, revealing the capability boundaries of current systems.

\section{Related Work}
\label{sec:related}

\begin{table*}[!t]
    \caption{\textbf{Comparison of world-model evaluation benchmarks.}
    We compare input formats, closed-loop adaptation, long-horizon revisitation, evaluated capabilities, and rollout duration.
    Objective denotes a scene-grounded high-level goal, while \cameraicon{} and \actionicon{} denote camera and action inputs.
    For \method, the Objective is accompanied by a basic action sequence that serves only as the Agent Player's initial reference; State Evolution includes visible and out-of-sight evolution.
    \checkmark, \ding{55}, and $\sim$ indicate full, no, and partial coverage, respectively.}
    \label{tab:criteria}
    \centering
    \renewcommand{\arraystretch}{1.18}
    \setlength{\tabcolsep}{3.5pt}
    \newcommand{\yes}{\textcolor{green!60!black}{\checkmark}}
    \newcommand{\no}{\textcolor{red}{\ding{55}}}
    \newcommand{\partyes}{\textcolor{orange!80!black}{$\sim$}}
    \resizebox{\linewidth}{!}{%
        \begin{tabular}{llcccccc}
            \toprule
            \textbf{Benchmark}
            & \textbf{Input}
            & \makecell{\textbf{Closed-Loop}\\\textbf{Adaptation}}
            & \makecell{\textbf{Long-horizon}\\\textbf{Revisit}}
            & \makecell{\textbf{Geometry}\\\textbf{Consistency}}
            & \makecell{\textbf{State}\\\textbf{Evolution}}
            & \makecell{\textbf{Interaction}\\\textbf{Fidelity}}
            & \makecell{\textbf{Time Scale}\\\textbf{/ Video}} \\
            \midrule
            WorldScore~\citep{worldscore}                                                 & Text + Image + \cameraicon{}                & \no      & \no  & \yes     & \no      & \no      & $\sim$2--10 s  \\
            WorldMark~\citep{xu2026worldmark}                                             & Image + \actionicon{}                       & \no      & \partyes & \yes     & \no      & \no      & 20 / 40 / 60 s \\
            WBench~\citep{ying2026wbench}                                                 & Text + Image + \cameraicon{}/\actionicon{}  & \no      & \partyes & \yes     & \partyes & \yes     & 10--45 s   \\
            WorldRoamBench~\citep{xu2026worldroambench}                                   & Image + \actionicon{}                       & \no      & \partyes & \yes     & \partyes & \yes     & 10--60 s        \\
            MemoBench~\citep{chen2026memobench}                                           & Text + Image + \cameraicon{}                & \no      & \no  & \yes     & \yes     & \no      & 4--6 s  \\
            Omni-WorldBench~\citep{wu_OmniWorldBenchComprehensiveInteractionCentric_2026} & Text + Image + \cameraicon{}                & \no      & \no  & \partyes & \yes     & \yes     & 3--6 s          \\
            \midrule
            \rowcolor{grouplavender}
            \method~(Ours)                                                               & Text + Image + Objective                     & \yes     & \yes & \yes     & \yes     & \yes     & 10--60 s \\
            \bottomrule
        \end{tabular}%
    }
    \vspace{-0.8\baselineskip}
\end{table*}

\noindent \textbf{World models.} Video world models are rapidly developing across diverse application domains~\citep{bruce_GenieGenerativeInteractive_2024,he_Matrixgame20Opensource_2025}.
In autonomous driving, world models predict how traffic scenes change in response to ego motion, vehicle controls, and surrounding agents.
GAIA-1~\citep{hu2023gaia1}, DriveDreamer~\citep{wang2023drivedreamer}, DrivingWorld~\citep{hu2024drivingworld}, and Vista~\citep{gao2024vista} generate future observations for scenario simulation, planning, and policy evaluation.
In robotics, world models instead predict future visual observations conditioned on actions, context, or task instructions.
IRASim~\citep{zhu2024irasim}, Cosmos~\citep{nvidia_CosmosWorldFoundation_2025}, RoboScape~\citep{shang2025roboscape}, and LVP~\citep{chen2025largevideoplanner} use these predictions to support robot learning, planning, and control.
Beyond these domain-specific applications, general-purpose interactive world models generate visual environments continuously in response to user actions rather than producing a fixed video from a text or image prompt.
Representative systems include Matrix-Game~\citep{he_Matrixgame20Opensource_2025,wang_MatrixGame30RealTime_2026}, YUME~\citep{yume15,yume10}, WorldPlay~\citep{worldplay2025}, LingBot-World~\citep{lingbot}, LingBot-World2~\citep{gao2026lingbotworld2}, SANA-WM~\citep{zhu2026sanawm}, HY-World~\citep{hyworld2025}, HY-World2~\citep{hyworld2026}, and Hunyuan-GameCraft~\citep{li_HunyuanGameCraftHighdynamicInteractive_2025}.
These models progressively generate the environment in response to movement keys, mouse or gamepad controls, camera trajectories, or text instructions.
Closed-source systems such as Genie 3~\citep{genie3} and HappyOyster~\citep{happyoyster} provide similar interaction through user-facing web interfaces, although their implementation details remain unavailable.
These systems use different architectures and control formats but share the goal of generating worlds that respond continuously and coherently to user actions.
Open-ended exploration and adventure scenarios place users in direct control of the generated environment.
We therefore consider them a fundamental setting.

\noindent\textbf{Evaluation of world models.}
Existing world-model benchmarks evaluate several complementary capabilities.
VBench~\citep{vbench} and T2V-CompBench~\citep{sun2025t2vcompbench} measure perceptual quality, temporal stability, and compositional alignment, but do not assess responses to user actions.
WorldSimBench~\citep{qin_WorldSimBenchVideoGeneration_2024} evaluates perceptual quality and downstream embodied-control performance, without adapting evaluation controls according to generated observations.
WorldScore~\citep{worldscore} and WorldMark~\citep{xu2026worldmark} assess camera control and multi-view geometry under predefined camera trajectories or action sequences.
Their results may conflate geometric inconsistency with trajectory failure when identical controls produce different movement magnitudes across models.
MIND~\citep{ye_MINDBenchmarkingMemory_2026}, MemoBench~\citep{chen2026memobench}, and STEVO-Bench~\citep{ma_OutSightOut_2026} evaluate memory and out-of-sight evolution through revisitation, visible--disappear--reappear processes, and lookaway controls.
These protocols follow fixed schedules that cannot account for model-specific action granularity and response speed.
Omni-WorldBench~\citep{wu_OmniWorldBenchComprehensiveInteractionCentric_2026} broadens evaluation to interaction, physics, and memory, while WBench~\citep{ying2026wbench} and WorldRoamBench~\citep{xu2026worldroambench} introduce predefined multi-turn navigation patterns.
However, these patterns cannot be flexibly composed into diverse long-horizon trajectories or adjusted online when the generated observation deviates from the expected state.
Overall, existing benchmarks predominantly drive world models with predefined low-level controls.
Because models differ in action granularity and response dynamics, the same controls may fail to reach comparable evaluation states.
\method specifies a shared long-horizon objective and uses a multi-modal Agent Player to adapt action execution online.
This closed-loop protocol preserves a consistent evaluation intent across models while supporting more diverse long-horizon interaction.

\section{\method}
\label{sec:worldplay}

To better simulate how human players evaluate world models, \method introduces a multi-modal Agent Player that interacts with each evaluated model and adaptively adjusts action execution.
The \method benchmark organizes long-horizon objectives across diverse scenario types.
Its VQA rubric verifier assesses four core world-model capabilities: geometry consistency, interaction fidelity, out-of-sight evolution, and insight evolution. We also combine basic quality metrics for complementary evaluation.

\subsection{Agent Player Design}
\label{sec:worldplay_player}

The Agent Player consists of a replaceable multi-modal agent model~(\eg, Claude or Gemini) and an agent interface.
The agent model observes generated frames and execution history, then decides how to adapt action execution toward the long-horizon objective.
The agent interface executes these decisions on the evaluated world model, captures the resulting observation, and returns it to the agent model, forming a closed interaction loop.

\noindent\textbf{Agent model decision.}
A straightforward way to instantiate the Agent Player is to let the agent plan the trajectory from scratch and interact with the world model without any human-provided action input.
However, because each decision requires the agent model to reason over the current observation, fully online planning introduces substantial decision latency and slows action execution.
It may also produce highly different trajectories across world models with different capabilities, making cross-model comparison less consistent.
We therefore use a human-annotated basic action sequence as a shared initial reference.
This preserves a consistent evaluation intent across models, reduces the burden of open-ended sequential planning, and still allows observation-conditioned adaptation.
As shown on the left of Fig.~\ref{fig:teaser}, the Agent Player receives the long-horizon objective and the shared basic action sequence.
At each interaction step, the agent model observes the generated frame, the previously executed actions, the scene description, and the objective.
Based on this context, it decides whether to keep, stop, extend, correct, or end the current rollout.
\textit{Keep} retains the planned action, continuing the current hold or advancing to the next action as scheduled.
\textit{Stop} terminates the active action early when the intended visual state has been reached.
\textit{Extend} increases the action hold duration when additional movement is required.
\textit{Correct} revises or skips the next planned action when the observed state no longer matches the reference sequence.
\textit{End} terminates the evaluation case only when the Agent Player determines that the long-horizon objective and its required observation have been completed.
For interaction fidelity cases involving an obstacle, reaching the obstacle does not trigger \textit{End}; the Agent Player continues the forward action so that the rollout reveals whether the subject collides with the obstacle realistically or penetrates it.

\noindent\textbf{Agent interface execution.}
The agent interface directly interacts with the evaluated world model.
It receives an initial frame and a human-annotated basic action sequence expressed using W/A/S/D, $\uparrow$/$\downarrow$/$\leftarrow$/$\rightarrow$, and WAIT, and translates these actions into the world model's native controls.
At each step, it executes the active action, captures the resulting generated frame, returns the frame to the agent model, and applies the returned decision to update subsequent execution.
This process continues in a closed loop until the Agent Player returns \textit{End} or the 40-step interaction budget is exhausted.
For web-based world models, the agent interface uses browser automation to dispatch controls and capture generated frames directly from the model interface.
Further details are provided in the supplementary.

\begin{figure*}[!t]
    \centering
    \includegraphics[width=\linewidth]{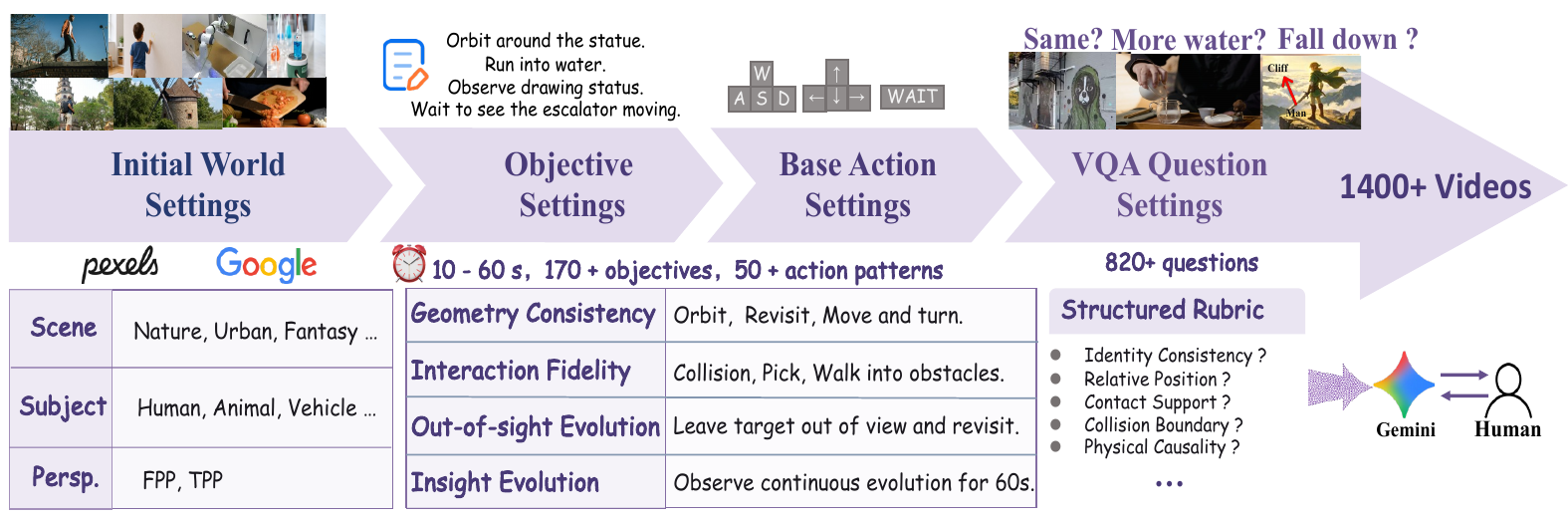}
    \vspace{-18pt}
    \caption{\textbf{Benchmark construction pipeline of \method.}
    Starting from diverse initial worlds, annotators define long-horizon objectives based on scenario-specific attributes, write basic action sequences, and construct sample-specific VQA rubrics.
    The resulting benchmark spans over 170 human-annotated cases and 50 action patterns over 10--60-second rollouts, with more than 820 questions applied to over 1,400 interactive videos.}
    \label{fig:construction}
\end{figure*}

\subsection{PlayWorld Benchmark Overview}
\label{sec:worldplay_suite}

\noindent\textbf{Benchmark construction.}
To support long-horizon  evaluation, we construct the \method benchmark around three core elements: an initial world, a scene-grounded objective, and a basic action sequence.
As shown in Fig.~\ref{fig:construction}, we first collect candidate images from Pexels~\citep{pexels} and Google Images~\citep{googleimages}, covering natural, urban, and fantasy environments with various subjects such as humans, animals, and vehicles.
Annotators then screen the images for clear scene structure and select first-person or third-person views according to the target capability and scene content.
For each selected image, Gemini~3.1 Pro~\citep{google2026gemini31pro} generates an environment caption, which is verified and revised by human annotators.
Annotators further define a long-horizon objective based on scenario-specific attributes, specify a visually verifiable completion condition, and write a basic action sequence.
Finally, annotators construct sample-specific VQA rubrics for the target evaluation dimension.
All cases are manually curated to ensure that the expected outcome is observable and suitable for dimension-specific evaluation.

\begin{figure*}[!t]
    \centering
    \includegraphics[width=\linewidth]{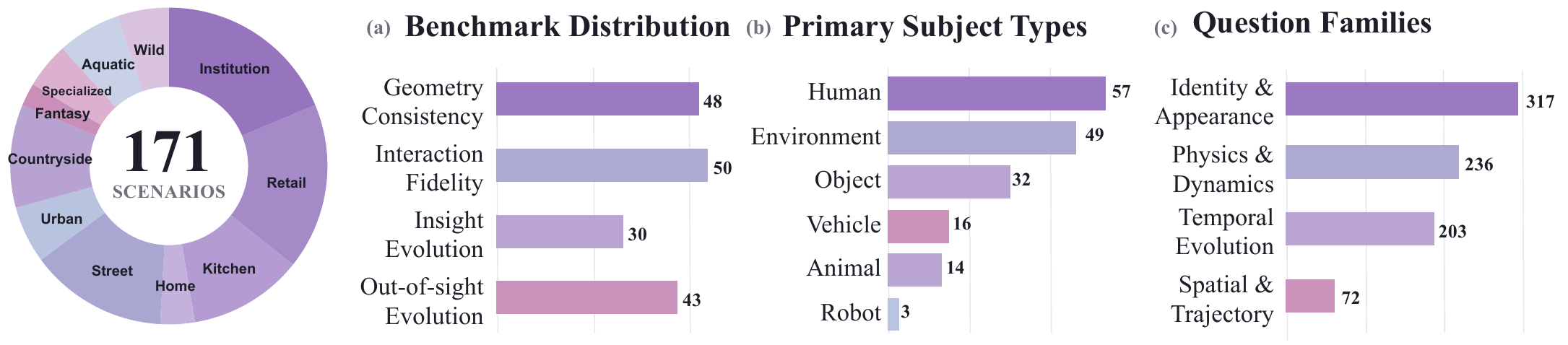}
    \vspace{-15pt}
    \caption{\textbf{Composition of \method.}
    The benchmark comprises over 170 human-annotated cases across diverse scenario categories.
    The cases span four evaluation dimensions and diverse subject types, with dimension-specific VQA questions tailored to their evaluation criteria.}
    \label{fig:dataset_statistics}
\end{figure*}

\begin{figure}[!t]
    \centering
    \includegraphics[width=0.99\linewidth]{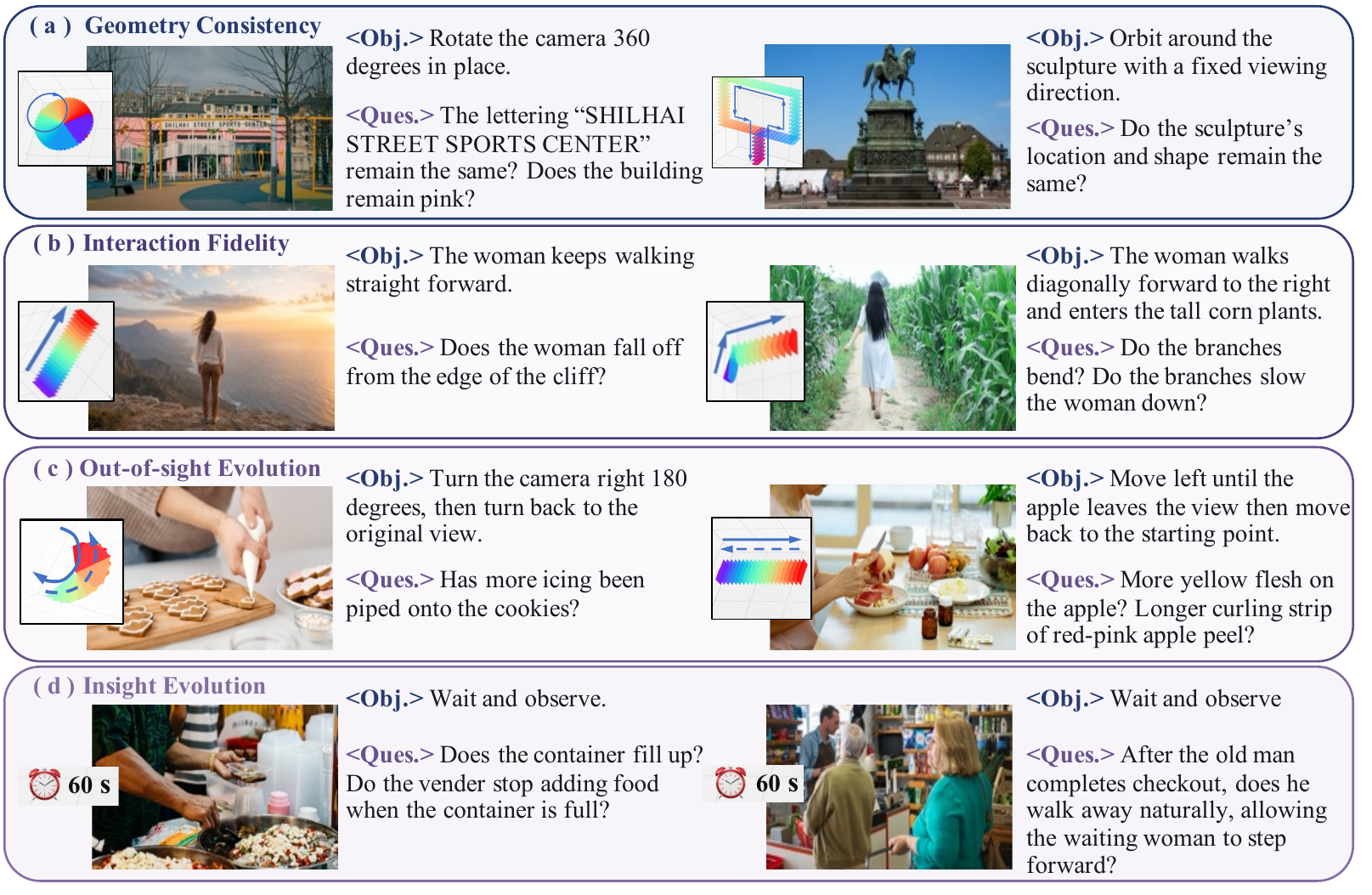}
    \vspace{-0.5\baselineskip}
    \caption{\textbf{Representative cases in \method.}
    Each case combines an initial world, a scene-grounded long-horizon objective, and a sample-specific VQA rubric.
    The expected trajectory is visualized alongside each case; for insight evolution, the camera remains stationary and observes the scene for 60 seconds.}
    \label{fig:suite_stats}
    \vspace{-0.6\baselineskip}
\end{figure}

\noindent\textbf{Benchmark composition.}
Fig.~\ref{fig:dataset_statistics} summarizes the composition of \method.
The benchmark comprises over 170 human-annotated cases covering 50 action patterns, ranging from a single action to combinations of up to five actions and producing 10--60-second rollouts.
Evaluating nine world models yields over 1,400 interactive videos, which are assessed using more than 820 task-conditioned VQA questions.
The cases span four evaluation dimensions and cover six primary subject types.
The VQA questions are organized into four evidence families: identity and appearance, physics and dynamics, temporal evolution, and spatial and trajectory reasoning.
Fig.~\ref{fig:suite_stats} provides representative examples for each dimension. Geometry consistency cases emphasize revisitable visual anchors such as lettering and buildings with distinctive shapes; interaction fidelity cases include walking into water, fire, cliffs, or grass; out-of-sight evolution cases emphasize continued changes across disappear--reappear intervals, such as a person peeling an apple, making dessert, or painting a picture; and insight evolution cases focus on long visible processes without an explicit action trajectory, such as shoppers queuing for checkout or a chef plating food.

\subsection{VQA Rubric Verifier}
\label{sec:worldplay_vqa}

Automated metrics primarily capture pixel-level fidelity and low-level perceptual quality, but often fail to measure consistency under large viewpoint changes, assess interaction plausibility, or determine whether long-horizon objectives are completed.
We therefore design a VQA rubric verifier, where Gemini~3.1~Pro~\citep{google2026gemini31pro} answers sample-specific Yes/No rubric questions for each generated rollout.
Each rubric assigns greater importance to questions that are more relevant to the target capability and excludes questions that are not applicable to the case.
Because instruction-following failures can make memory evaluation unreliable, as discussed in MemoBench~\citep{chen2026memobench}, the verifier uses dimension-specific validation as a gate before rubric scoring.
The weighted Yes/No outcomes are aggregated and converted into a case-level score on a 1--5 scale ($\uparrow$: higher is better), as reported in Tab.~\ref{tab:results}.
The evaluation covers four dimensions:

\noindent\textbf{Geometry consistency.}
To quantitatively assess long-horizon geometry consistency, the rubric considers whether the environment preserves its structure during camera movement and revisitation.
We first verify \textit{Trajectory Validity} by determining whether the rollout follows the intended path and reaches the required viewpoint.
Once the rollout passes this validation, its sample-specific VQA rubric covers two aspects.
\textit{Scene Identity} checks the preservation of object identity, count, texture, material, and color, such as whether a wall mural preserves its color and shape after a 360-degree turn or whether a revisited sculpture remains unchanged.
\textit{Spatial Consistency} considers relative object positions and the overall scene layout, for example whether a red car that is initially to the right of a green car appears in the correct left-right order after the camera crosses the street and looks back.

\noindent \textbf{Interaction fidelity.}
For interaction fidelity, we examine whether the intended interaction occurs and elicits a physically plausible response.
Before rubric scoring, \textit{Subject and Reachability} determines whether the controlled subject responds to the actions and reaches the intended interaction region.
A completely static subject fails this validation because the intended interaction is never tested.
The subsequent rubric covers \textit{Contact and Collision}, \textit{Motion and Causality}, and \textit{Visual Response}.
For instance, it asks whether the subject collides with a wall rather than passing through it and whether its motion follows plausible kinematics, such as maintaining grounded footsteps without floating.
It also checks whether entering water produces appropriate physical feedback, such as visible ripples, splashes, or changes in apparent water depth around the subject.

\noindent\textbf{Out-of-sight evolution.}
Out-of-sight evolution focuses on whether a target preserves its identity and continues to change while unobserved.
We first use \textit{Trajectory Validity} to determine whether the target leaves the field of view and subsequently reappears as required.
For eligible rollouts, \textit{Reappearance Consistency} asks whether the revealed target is the same entity observed before it disappeared.
\textit{Hidden-State Evolution} determines whether the target state progresses plausibly during the unobserved interval, such as whether an in-progress drawing gains additional content while it is out of sight.
\textit{Physical Causality} checks whether the revealed state can reasonably result from that hidden evolution, such as a person continuing to move upward on an escalator, blue ink spreading through clear water, or a cup becoming fuller as water is poured into it.

\noindent\textbf{Insight evolution.}
Insight evolution focuses on continuously observed phenomena without a fixed action trajectory.
The rubric evaluates a visible subject or process from three aspects.
\textit{Identity and State Progression} checks whether the relevant person or object remains present and whether its state progresses over time, rather than suddenly disappearing.
\textit{Motion and Physical Plausibility} identifies static behavior, repetitive but non-progressive actions, abrupt transitions, and implausible dynamics.
For example, if a person is eating one of three buns, the rubric checks whether the bun being eaten gradually changes and whether the number or appearance of the remaining buns evolves consistently.
Meanwhile, \textit{Temporal Scene Consistency} detects unintended changes to the surrounding environment.
Because this dimension uses stationary observation, it does not require trajectory validation.

\begin{table*}[!t]
    \caption{\textbf{Rubric-based VQA evaluation across four dimensions on the \method benchmark.}
    Scores range from 1 to 5 ($\uparrow$: higher is better).
    Overall is the unweighted mean across the four evaluation dimensions.
    \textbf{Bold}: best; \underline{underline}: second best.}
    \label{tab:results}
    \centering
    \fontsize{8.5}{12}\selectfont
    \renewcommand{\arraystretch}{1.18}
    \setlength{\tabcolsep}{5pt}
    \begin{tabular}{lccccc}
        \toprule
        Model
            & \makecell{Geometry\\Consistency}
            & \makecell{Interaction\\Fidelity}
            & \makecell{Insight\\Evolution}
            & \makecell{Out-of-sight\\Evolution}
            & Overall
            \\
        \midrule
        \rowcolor{grouplavender}
        \multicolumn{6}{@{}l}{\emph{Closed-source Models}} \\
        \midrule
        Genie 3 & \textbf{2.74} & \textbf{2.40} & \underline{1.51} & \textbf{1.81} & \textbf{2.12} \\
        HappyOyster & \underline{2.54} & 2.15 & 1.47 & \underline{1.54} & \underline{1.92} \\
        \midrule
        \rowcolor{grouplavender}
        \multicolumn{6}{@{}l}{\emph{Open-source Models}} \\
        \midrule
        LingBot-World & 2.11 & \underline{2.23} & 1.33 & 1.43 & 1.78 \\
        LingBot-World2 & 2.04 & 2.13 & \textbf{1.95} & 1.16 & 1.82 \\
        HY-World2 & 2.14 & 2.06 & 1.13 & 1.09 & 1.61 \\
        SANA-WM & 1.72 & 1.89 & 1.13 & 1.16 & 1.48 \\
        Hunyuan-GameCraft & 1.62 & 1.52 & 1.21 & 1.31 & 1.42 \\
        HY-WorldPlay & 1.12 & 1.63 & 1.01 & 1.08 & 1.21 \\
        Matrix-Game-3.0 & 1.30 & 1.25 & 1.00 & 1.00 & 1.14 \\
        \bottomrule
    \end{tabular}
\end{table*}

\begin{table*}[!t]
    \caption{\textbf{Trajectory-validation pass rates on the \method benchmark.}
    The insight evolution dimension uses stationary observation and therefore does not require trajectory validation and is marked as ``--''.
    Overall is the unweighted mean of the pass rates for geometry consistency, interaction fidelity, and out-of-sight evolution.}
    \label{tab:pass_rates}
    \centering
    \fontsize{8.5}{12}\selectfont
    \renewcommand{\arraystretch}{1.18}
    \setlength{\tabcolsep}{5pt}
    \begin{tabular}{lccccc}
        \toprule
        Model
            & \makecell{Geometry\\Consistency}
            & \makecell{Interaction\\Fidelity}
            & \makecell{Insight\\Evolution}
            & \makecell{Out-of-sight\\Evolution}
            & Overall \\
        \midrule
        \rowcolor{grouplavender}
        \multicolumn{6}{@{}l}{\emph{Closed-source Models}} \\
        \midrule
        Genie 3 & \textbf{77.1\%} & \textbf{93.5\%} & -- & \underline{90.7\%} & \textbf{87.1\%} \\
        HappyOyster & \underline{63.8\%} & \textbf{93.5\%} & -- & 81.4\% & 79.6\% \\
        \midrule
        \rowcolor{grouplavender}
        \multicolumn{6}{@{}l}{\emph{Open-source Models}} \\
        \midrule
        LingBot-World & 58.3\% & 76.0\% & -- & 83.7\% & 72.7\% \\
        LingBot-World2 & 58.3\% & 85.1\% & -- & \textbf{93.0\%} & 78.8\% \\
        HY-World2 & 47.9\% & 70.0\% & -- & 37.2\% & 51.7\% \\
        SANA-WM & 62.5\% & \underline{88.0\%} & -- & \underline{90.7\%} & \underline{80.4\%} \\
        Hunyuan-GameCraft & 25.0\% & 78.0\% & -- & 81.4\% & 61.5\% \\
        HY-WorldPlay & 14.6\% & 80.0\% & -- & 30.2\% & 41.6\% \\
        Matrix-Game-3.0 & 37.5\% & \underline{88.0\%} & -- & 79.1\% & 68.2\% \\
        \bottomrule
    \end{tabular}
\end{table*}

\subsection{Basic Ability Evaluation}
\label{sec:worldplay_eval}

We further use automatic metrics to test video quality and controllability.
We choose Aesthetic Quality, Imaging Quality, Motion Smoothness, and Temporal Flickering from VBench~\citep{vbench}, Temporal Consistency from Omni-WorldBench~\citep{wu_OmniWorldBenchComprehensiveInteractionCentric_2026}, Depth Stability from MemoBench~\citep{chen2026memobench}, and Subject Consistency from HyDRA~\citep{hydra}.
These video quality metrics do not require ground-truth videos and mainly capture frame-level or inter-frame signals.
All video quality metrics are normalized to $[0,1]$ and reported as percentages.
We then evaluate Action Controllability using Translation Pass Rate and Rotation Pass Rate, following WorldMark~\citep{xu2026worldmark}.
Camera poses are estimated by VGGT~\citep{VGGT} from uniformly sampled frames, while target trajectories are defined by the executed actions, including online adjustments made by the Agent Player.
A rollout passes the translation or rotation criterion when the corresponding error falls below an acceptable threshold.
Both pass rates are computed over rollouts with valid estimates and reported as percentages, where higher values indicate better controllability.
Tab.~\ref{tab:automatic_results} reports all nine higher-is-better metrics together with a Basic Ability Score, which converts their average rank into a percentage.

\section{Experiments}
\label{sec:experiments}
\subsection{Models and Evaluation Protocol}
\noindent\textbf{Evaluation models.}
We evaluate nine representative world models.
Five models are evaluated through their web-based interfaces to reproduce the user-facing interaction setting: Genie 3~\citep{genie3}, LingBot-World~\citep{lingbot}, LingBot-World2~\citep{gao2026lingbotworld2}, HY-World2~\citep{hyworld2026}, and HappyOyster~\citep{happyoyster}.
Four models are executed locally through chunk-wise generation: SANA-WM~\citep{zhu2026sanawm}, Hunyuan-GameCraft~\citep{li_HunyuanGameCraftHighdynamicInteractive_2025}, HY-WorldPlay~\citep{worldplay2025}, and Matrix-Game-3.0~\citep{wang_MatrixGame30RealTime_2026}.
All models are controlled by the same \player, whose agent interface converts the decisions into each model's native input format.
Implementation details and generation configurations are provided in the supplementary material.

\begin{figure}[!t]
    \centering
    \includegraphics[width=0.99\linewidth]{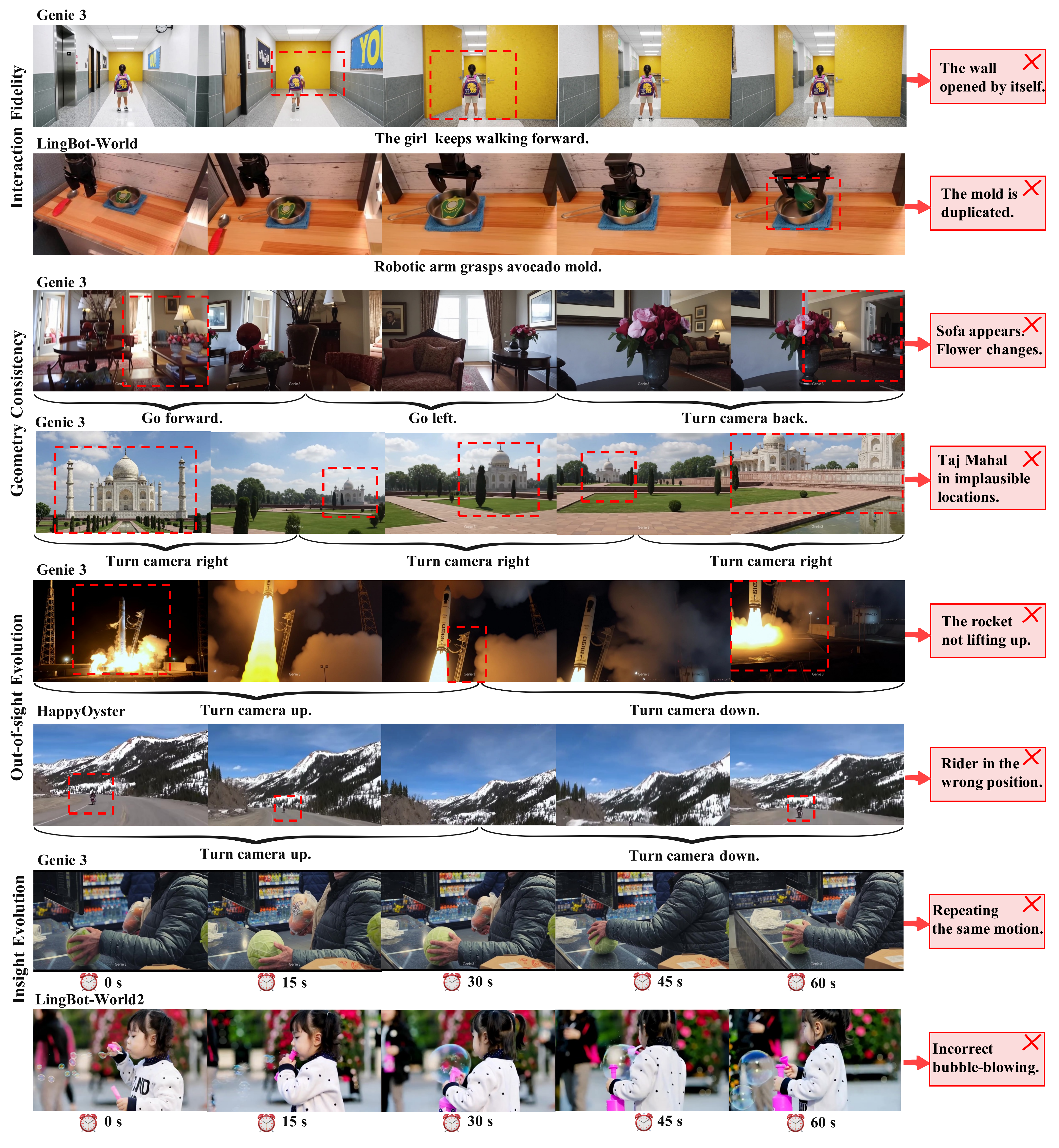}
     \vspace{-15pt}
     \caption{\textbf{Representative failures across four evaluation dimensions.}
    Current video world models can follow camera or action controls, but still struggle to simulate coherent and realistic world dynamics.}
    \label{fig:qualitative}
\end{figure}

\noindent\textbf{Evaluation protocol.}
For each evaluation case, all models receive the same initial world, long-horizon objective, and basic action sequence.
The same Agent Player observes the generated frames and adjusts action execution for every model.
We evaluate the resulting rollouts, which last approximately 10--60 seconds, using the rubric verifier in Sec.~\ref{sec:worldplay_vqa} and the basic ability metrics in Sec.~\ref{sec:worldplay_eval}.
The rubric verifier reports geometry consistency, interaction fidelity, out-of-sight evolution, and insight evolution, while the basic ability evaluation reports Video Quality and Action Controllability.
We also report dimension-specific validation pass rates in Tab.~\ref{tab:pass_rates}.
Specifically, geometry consistency is validated using Trajectory Validity, interaction fidelity using Subject and Reachability, and out-of-sight evolution using Trajectory Validity.
These three validation criteria are evaluated independently.
If a rollout does not pass the corresponding validation criterion, its score for that dimension is set to the minimum value of 1.
The insight evolution dimension uses stationary observation and therefore does not require trajectory validation.
\subsection{Agent Player Analysis}
\noindent\textbf{Control strategy.}
Existing benchmarks typically execute a predefined trajectory for each evaluation case.
We retain a basic action sequence as a shared initial reference but adapt its execution online to account for differences in action granularity across world models.
Directly executing the basic action sequence without online adaptation (Preset Only) may cause the same controls to overshoot or undershoot the intended state.
Agent Only instead predicts every action online from the latest generated observation and previously executed actions, but planning each action from scratch increases the sequential inference burden and slows trajectory execution.
Preset + Agent combines both sources of guidance: it uses the basic action sequence as a reference while adapting its execution to the generated observations and model-specific control granularity.
We compare these three strategies on the same 25 cases for Genie 3 and HappyOyster.

As shown in Tab.~\ref{tab:player_ablation}, Preset + Agent achieves the highest Trajectory Score and Human Preference on both world models, confirming the benefit of combining a shared action prior with online adaptation.
Its Agent-modified Action Ratio remains within $10$--$20\%$ on both models, showing that the Agent makes targeted modifications.
By adapting action execution online, \method reaches the intended objective more reliably than executing a predefined trajectory alone.

\noindent\textbf{Agent model selection.}
We further investigate the effect of multi-modal model choice on the trajectories produced by the Agent Player.
We compare three agent models on the same Genie 3 cases using \method's control strategy.
As shown in Tab.~\ref{tab:agent_model_selection}, agent model choice produces only minor differences in trajectory quality.

We attribute this limited sensitivity to the relatively constrained decision task, in which each agent model adapts actions based on the current visual state and basic action sequence rather than planning an open-ended trajectory from scratch.
Claude Sonnet achieves the highest Trajectory Score, whereas Claude Haiku achieves the highest Human Preference and the lowest measured Decision Latency.
Because latency depends strongly on provider infrastructure and network conditions, the measured values may vary across deployment environments.
Considering the overall quality--efficiency trade-off, we use Claude Haiku, and the agent model can be readily upgraded as stronger multi-modal models emerge.

\noindent\textbf{Implementation details.}
For each comparison, we aggregate judgments from five independent raters with video-generation experience.
Trajectory Score is assigned by Gemini~3.1~Pro on a 0--2 scale.
Human Preference aggregates explicit pairwise preference judgments, assigning full credit to wins and half credit to ties.
For the control-strategy comparison, Preset Only and Agent Only are each compared against Preset + Agent, whose rate pools comparisons against both baselines.
For the agent model comparison, each agent model pools comparisons against the other agent models.
The comparison order and left--right placement of the paired trajectories are randomized in the questionnaire.
Majority Agreement is the percentage of pairwise comparisons for which more than half of the raters select the same outcome.
Agent-modified Action Ratio is the macro-averaged per-case fraction of actions modified or inserted by the Agent Player.
Decision Latency, reported only for the agent model comparison, is the average end-to-end waiting time per agent-model call.

\begin{table*}[!t]
    \caption{\textbf{Comparison of trajectory control strategies.}
    Preset Only executes the basic action sequence without online adaptation.
    Agent Only selects all actions online from generated observations and executed-action history.
    Preset + Agent adapts action execution online using the basic action sequence as a reference.}
    \label{tab:player_ablation}
    \centering
    \fontsize{8.5}{12}\selectfont
    \renewcommand{\arraystretch}{1.18}
    \setlength{\tabcolsep}{7pt}
    \begin{tabular}{lcccc}
        \toprule
        Setting
            & \makecell{Trajectory\\Score $\uparrow$}
            & \makecell{Human\\Preference $\uparrow$}
            & \makecell{Majority\\Agreement}
            & \makecell{Agent-modified\\Action Ratio} \\
        \midrule
        \rowcolor{grouplavender}
        \multicolumn{5}{@{}l}{\emph{Genie 3}} \\
        \midrule
        Preset Only & 0.92 & 39.6\% & 88.0\% & 0.0\% \\
        Agent Only & 0.88 & 29.2\% & 84.0\% & 100.0\% \\
        Preset + Agent & \textbf{1.08} & \textbf{65.6\%} & 86.0\% & 12.0\% \\
        \midrule
        \rowcolor{grouplavender}
        \multicolumn{5}{@{}l}{\emph{HappyOyster}} \\
        \midrule
        Preset Only & 1.00 & 40.8\% & 76.0\% & 0.0\% \\
        Agent Only & 0.68 & 24.4\% & 92.0\% & 100.0\% \\
        Preset + Agent & \textbf{1.12} & \textbf{67.4\%} & 84.0\% & 14.9\% \\
        \bottomrule
    \end{tabular}
\end{table*}

\begin{table*}[!t]
    \caption{\textbf{Comparison of agent models.}
    We conduct the comparison on Genie 3, with Preset + Agent control strategy.
    We recommend Claude Haiku as the agent model.}
    \label{tab:agent_model_selection}
    \centering
    \fontsize{8.5}{12}\selectfont
    \renewcommand{\arraystretch}{1.18}
    \setlength{\tabcolsep}{3pt}
    \begin{tabular}{lccccc}
        \toprule
        \makecell{Agent\\Model}
            & \makecell{Trajectory\\Score $\uparrow$}
            & \makecell{Human\\Preference $\uparrow$}
            & \makecell{Majority\\Agreement}
            & \makecell{Agent-modified Action\\Ratio}
            & \makecell{Decision Latency\\(s/call) $\downarrow$} \\
        \midrule
        \rowcolor{grouplavender}
        Claude Haiku & 1.08 & \textbf{57.8\%} & 88.9\% & 12.0\% & \textbf{3.83} \\
        Claude Sonnet & \textbf{1.24} & 45.3\% & 94.1\% & 12.4\% & 6.21 \\
        Gemini 3.1 Pro & 1.08 & 46.5\% & 94.1\% & 12.6\% & 4.36 \\
        \bottomrule
    \end{tabular}
\end{table*}

\subsection{Observations in World Models}

\noindent\textbf{Sustained world evolution remains the primary bottleneck.}
Tab.~\ref{tab:results} shows that Genie 3 achieves the highest geometry consistency, interaction fidelity, out-of-sight evolution, and overall scores, while LingBot-World2 leads insight evolution and HappyOyster ranks second overall.
However, out-of-sight evolution and insight evolution consistently receive lower scores across models.
This suggests that current world models are better at preserving visible structure or producing immediate responses than sustaining semantic state changes over long horizons.
In practice, during continuous observation, an ongoing process may remain static, reset, or evolve without preserving its causal order.
After leaving the field of view, a target may reappear unchanged, return at an incorrect location, or be replaced by another entity.
Together, these results indicate that persistent state evolution across time and occlusion remains a central challenge for current world models.

\noindent\textbf{Long-horizon revisitation exposes global spatial inconsistency.}
Fig.~\ref{fig:qualitative} shows that failures that appear minor in individual frames can accumulate into structural inconsistencies over a complete rollout.
Across multiple orbit and revisit cases, salient landmarks are repeatedly generated at new viewpoints rather than anchored to unique spatial locations.
For example, when the camera orbits 360 degrees around the Taj Mahal, the model may repeatedly regenerate the monument at different viewpoints, creating multiple inconsistent copies of the same landmark.
Although each frame remains visually plausible, the complete sequence violates spatial uniqueness and global scene topology.
This failure is consistent with current models relying heavily on local appearance continuity and short-range motion cues, without reliably maintaining a persistent global representation of the generated 3D world.

\noindent\textbf{Interaction fidelity remains limited beyond simple collision responses.}
Current models handle simple interactions reasonably well.
In third-person cases, several leading models can stop a character at an obstacle boundary, whereas first-person movement may still pass directly through solid obstacles, as shown in Fig.~\ref{fig:qualitative}.
More complex interactions, such as walking into water, often fail to produce the expected physical or visual response.
In addition, third-person models do not consistently bind input actions to the controlled character, resulting in delayed, insufficient, or unrelated motion.

\begin{table*}[!t]
    \caption{\textbf{Basic ability of video world models on the \method benchmark.}
    Video Quality includes Aesthetic Quality, Imaging Quality, Motion Smoothness, Temporal Flickering, Temporal Consistency, Depth Stability, and Subject Consistency.
    Action Controllability includes Translation Pass Rate and Rotation Pass Rate.
    Each pass rate is computed over rollouts with a valid estimate.
    Basic Ability Score converts the average rank across all nine metrics into a higher-is-better percentage.
    All values are reported as percentages (\%).
    \textbf{Bold}: best; \underline{underline}: second best.}
    \label{tab:automatic_results}
    \centering
    \renewcommand{\arraystretch}{1.18}
    \resizebox{\linewidth}{!}{%
    \begin{tabular}{lcccccccccc}
        \toprule
        \multirow{2}{*}{Model}
            & \multicolumn{7}{c}{Video Quality}
            & \multicolumn{2}{c}{Action Controllability}
            & \multirow{2}{*}{\makecell{Basic Ability\\Score $\uparrow$}} \\
        \cmidrule(lr){2-8} \cmidrule(lr){9-10}
            & Aes. $\uparrow$ & Img. $\uparrow$ & Mot. $\uparrow$ & Flick. $\uparrow$ & Temp. $\uparrow$
            & Depth $\uparrow$
            & Subj. $\uparrow$
            & Translation $\uparrow$
            & Rotation $\uparrow$
            & \\
        \midrule
        \rowcolor{grouplavender}
        \multicolumn{11}{@{}l}{\emph{Closed-source Models}} \\
        \midrule
        Genie 3 & \textbf{52.00} & \textbf{75.22} & 99.00 & 97.79 & \underline{98.64} & 88.70 & 85.60 & 64.1 & \textbf{50.6} & \underline{72.2} \\
        HappyOyster & 49.66 & 73.66 & \textbf{99.46} & \underline{99.02} & \textbf{99.52} & \underline{91.80} & \underline{88.40} & 58.0 & \underline{48.5} & \textbf{76.4} \\
        \midrule
        \rowcolor{grouplavender}
        \multicolumn{11}{@{}l}{\emph{Open-source Models}} \\
        \midrule
        LingBot-World & 51.12 & 72.04 & 97.96 & 96.27 & 98.02 & \textbf{92.20} & 83.50 & 64.8 & 44.7 & 51.4 \\
        LingBot-World2 & 51.47 & \underline{73.98} & 97.84 & 95.64 & 96.88 & 91.40 & 82.30 & 67.9 & 45.6 & 50.0 \\
        HY-World2 & 47.43 & 67.79 & \underline{99.12} & \textbf{99.46} & 91.17 & 90.70 & \textbf{89.80} & 40.8 & 42.2 & 45.8 \\
        SANA-WM & \underline{51.75} & 72.59 & 98.92 & 96.48 & 98.20 & 84.40 & 85.90 & \underline{78.6} & 42.2 & 56.9 \\
        Hunyuan-GameCraft & 50.14 & 67.82 & 98.25 & 95.24 & 97.97 & 90.40 & 83.70 & \textbf{89.6} & 29.4 & 40.3 \\
        HY-WorldPlay & 47.71 & 61.53 & 98.89 & 95.18 & 95.45 & 91.00 & 87.40 & 75.8 & 44.2 & 38.9 \\
        Matrix-Game-3.0 & 44.10 & 66.03 & 98.90 & 96.92 & 95.23 & 87.30 & 65.70 & 42.4 & 27.7 & 18.1 \\
        \bottomrule
    \end{tabular}}
\end{table*}

\noindent\textbf{Trajectory control and world-model capability can diverge.}
Tab.~\ref{tab:pass_rates} reports whether rollouts follow the trajectories required by the objectives.
The results show that strong trajectory control does not necessarily imply strong world-model ability.
For example, SANA-WM obtains a relatively high overall validation pass rate, suggesting that its trajectory control often reaches the intended objective region.
Nevertheless, its rubric scores remain modest.
This gap indicates that SANA-WM can often follow the required trajectory, but still struggles to preserve memory, maintain spatial consistency, or generate physically plausible interactions.

\noindent\textbf{Automatic metrics only test the basic ability of interactive world models.}
As shown in Tab.~\ref{tab:automatic_results}, most evaluated models perform well on Video Quality, while their Action Controllability remains less precise.
Moreover, passing the action-control thresholds does not guarantee that a rollout reaches the objective-specific target state or completes its long-horizon objective.
Depth Stability and Subject Consistency may also remain high when a model produces little or no camera motion, even though the intended evaluation objective is not completed.
Although HappyOyster achieves the highest Basic Ability Score, this does not imply that it has the strongest long-horizon world-model capability.

\begin{figure*}[!t]
    \centering
    \vspace{-15pt}
\includegraphics[width=\linewidth]{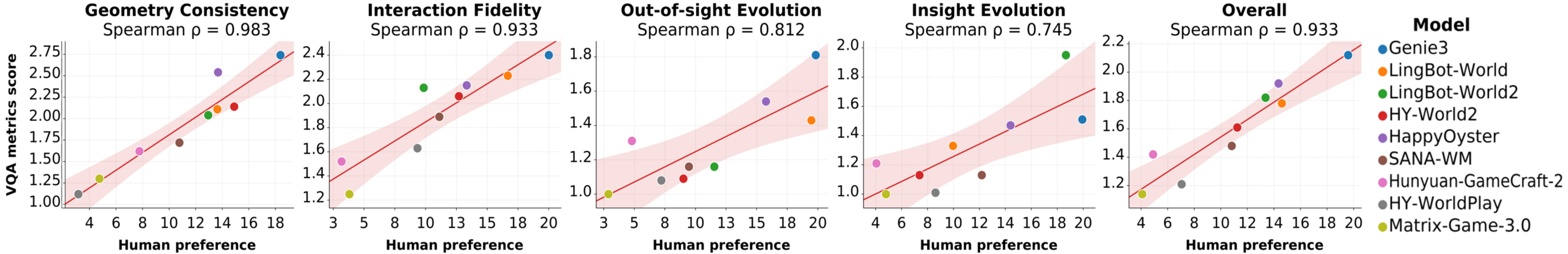}
 \vspace{-15pt}
    \caption{Alignment between human preferences and VQA metrics.}
    \label{fig:human_alignment}
\end{figure*}

\subsection{Human Validation}
\label{sec:human_evaluation}

We conduct a human evaluation to examine whether the VQA metrics align with human perception.
Participants provide 600 valid pairwise judgments on video pairs evenly sampled across the four dimensions.
We aggregate pairwise wins for each model within each dimension and normalize the resulting preference scores.
Each point in Fig.~\ref{fig:human_alignment} represents an evaluated model; the red line shows a linear fit, and the shaded region denotes its confidence band.
Spearman's $\rho$ measures rank agreement across the nine models for each evaluation dimension and the Overall score.
The consistently positive correlations show that the VQA metrics closely preserve human preference rankings.
Further details are provided in Appendix~\ref{sec:supp_human_annotation}.

\section{Conclusion}
\label{sec:conclusion}

We have presented \method, a benchmark that uses multi-modal Agent Players to simulate how human users evaluate interactive video world models.
With 171 human-annotated cases and an end-to-end automated evaluation system, \method supports consistent evaluation under shared long-horizon objectives.
Together, \method establishes a practical basis for more precise evaluation of video world models.

\section*{Acknowledgments}
We thank Jing He, Zhekai Chen, Wenwang Huang, Junjie Wang, Xianzhe Fan, Chenhao Ye, Tianxing Wu, and Junwei Luo for fruitful discussions.

\bibliographystyle{iclr2027_conference}
\bibliography{sample}

\newpage
\appendix

\section{Implementation Details}
\label{sec:supp_protocol}

\noindent\textbf{Closed-loop execution.}
Every evaluation case provides all models with the same initial frame, long-horizon objective, and human-annotated basic action sequence.
The agent interface executes the current action and returns the latest generated frame to the Agent Player, which observes the frame, the remaining basic action sequence, and the recent action history before returning \textit{Keep}, \textit{Stop}, \textit{Extend}, \textit{Correct}, or \textit{End}.
This closed loop continues until the Agent Player returns \textit{End} or the 40-step interaction budget is exhausted, producing rollouts of approximately 10--60 seconds.
We use Claude Haiku~4.5~\citep{anthropic2025claudehaiku45} as the agent model for all reported benchmark experiments.

\noindent\textbf{Model execution.}
Genie 3, LingBot-World, LingBot-World2, HY-World2, and HappyOyster are evaluated through their user-facing web interfaces, whereas SANA-WM, Hunyuan-GameCraft, HY-WorldPlay, and Matrix-Game-3.0 are evaluated locally through chunk-wise generation and model-specific adapters.
All settings use the same action definitions and record the executed actions and generated observations.
HY-World2 requires an input image that supports global scene modeling.
Cases whose initial images do not satisfy this requirement are skipped and excluded from its score aggregation.
Hunyuan-GameCraft requires a non-empty action input for every generated chunk; consequently, insight evolution cases that require prolonged passive observation or WAIT-only evolution cannot always be evaluated under their intended protocol.

\noindent\textbf{Basic ability metrics.}
Video Quality includes seven higher-is-better metrics.
Aesthetic Quality, Imaging Quality, Motion Smoothness, and Temporal Flickering follow VBench~\citep{vbench}; Temporal Consistency follows Omni-WorldBench~\citep{wu_OmniWorldBenchComprehensiveInteractionCentric_2026}; Depth Stability uses Depth Anything V2~\citep{yang2024depthanythingv2} following MemoBench~\citep{chen2026memobench}; and Subject Consistency is adapted from HyDRA's $\mathrm{DSC}_{\mathrm{ctx}}$~\citep{hydra}.
Subject Consistency detects dynamic subjects with YOLO~\citep{YOLO}, extracts CLIP~\citep{CLIP} features from the subject crops, and measures their cross-window similarity.
When no corresponding dynamic subject is detected in both windows, Subject Consistency is reported as N/A.

For Action Controllability, VGGT~\citep{VGGT} estimates camera poses from uniformly sampled frames following WorldMark~\citep{xu2026worldmark}.
The executed action sequence, including Agent Player adjustments, defines the target trajectory.
A rollout passes Translation when its target-normalized Translation Error is below $0.3$ and passes Rotation when its mean geodesic Rotation Error is below $45^\circ$.
Each pass rate is computed only over rollouts with a valid estimate for the corresponding metric.
Basic Ability Score ranks models independently on the seven Video Quality metrics and two Action Controllability pass rates, averages the nine ranks, and converts the result into a higher-is-better percentage.

\noindent\textbf{VQA rubric verifier.}
Gemini~3.1 Pro~\citep{google2026gemini31pro} serves as the VQA rubric verifier.
For geometry consistency, interaction fidelity, and out-of-sight evolution, dimension-specific validation first checks whether the rollout follows the trajectory required by the objective.
Only valid rollouts proceed to rubric scoring; invalid rollouts receive the minimum dimension score of 1.
Insight evolution uses stationary observation and therefore does not require trajectory validation.

To balance temporal coverage and spatial detail, the verifier receives two contact-sheet streams.
The primary stream samples the rollout at 10 FPS, resizes each frame to $384\times216$, and groups every 25 frames into a $5\times5$ grid.
The detail stream samples at 0.5 FPS, resizes each frame to $800\times450$, and groups every four frames into a $2\times2$ grid.
Gemini receives both streams together with the case objective and its sample-specific VQA rubric.
For each applicable question, it returns a binary answer $a_j\in\{0,1\}$ and supporting evidence.
Regular questions receive weight $1$, the question expressing the defining requirement receives weight $2$, and N/A questions receive weight $0$.
The dimension score is
\[
R_d=\frac{\sum_{j\in A_d}w_ja_j}{\sum_{j\in A_d}w_j},
\qquad
S_d=1+4R_d,
\]
where $A_d$ contains only applicable questions.
The Overall score is the unweighted mean of geometry consistency, interaction fidelity, out-of-sight evolution, and insight evolution.

\noindent\textbf{Radar-chart scales.}
In Fig.~\ref{fig:teaser}, the four rubric-based dimensions---geometry consistency, interaction fidelity, out-of-sight evolution, and insight evolution---are displayed on a range of $1$--$3$, while Basic Ability is displayed on a range of $0$--$1$.
The axes therefore visualize performance within their respective ranges and should not be interpreted as sharing the same absolute scale.

\section{VQA Scoring Robustness}
\label{sec:supp_vqa_robustness}

Gemini's outputs may vary across repeated calls even when the prompt and visual inputs are fixed.
Repeating rubric verification for every rollout is costly at the full benchmark scale, so benchmark evaluations commonly rely on a single scoring pass under a fixed configuration.
We likewise use one VQA scoring pass per rollout for the reported results and conduct an additional independent pass to assess robustness.
Across the nine models, the mean per-model sample variance between the two scoring passes is $0.0112$.
This low variance indicates that a single fixed scoring pass provides stable aggregate results, although the two-pass analysis is not a precise estimate of scoring uncertainty.
When the evaluation budget permits, we recommend averaging multiple independent VQA scoring passes to further reduce verifier variance.

\section{Agent Interface for Web-Based Models}
\label{sec:supp_playwright}

Web-based world models do not expose a common local inference API.
The agent interface therefore connects to an authenticated browser session, initializes the evaluation case, dispatches model-specific controls, monitors generation, and captures the resulting observation.
The captured frame is returned to the Agent Player, and the interface executes the next control after \textit{Keep}, \textit{Stop}, \textit{Extend}, or \textit{Correct}, or terminates the episode after \textit{End}.
For reproducibility, it records the task identifier, initial frame, basic action sequence, executed controls, intermediate frames, Agent Player decisions, timing, termination condition, and final video.

\noindent\textbf{Browser interaction.}
For each case, the agent interface opens the model interface, supplies the initial-world condition, selects the required perspective when this option is available, and waits until the generated world becomes interactive.
During execution, controls are dispatched through the browser interface and the rendered world is captured after each step.
The Agent Player decides only how the world should be controlled, while login, page navigation, submission, rendering-state checks, and output collection remain deterministic interface operations.

\begin{figure}[t]
    \centering
    \includegraphics[width=0.32\linewidth]{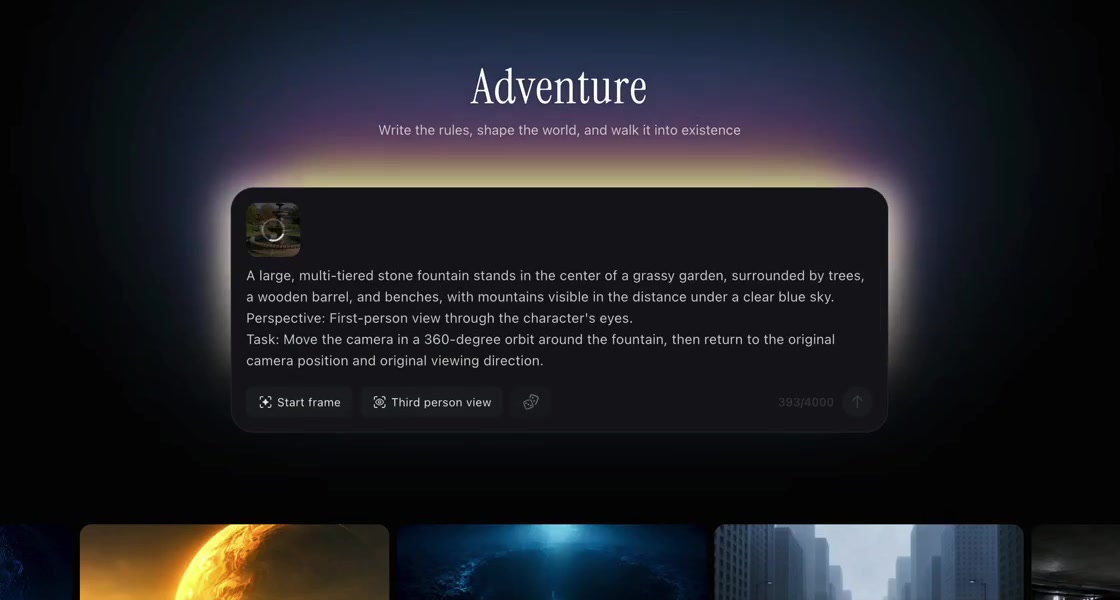}
    \includegraphics[width=0.32\linewidth]{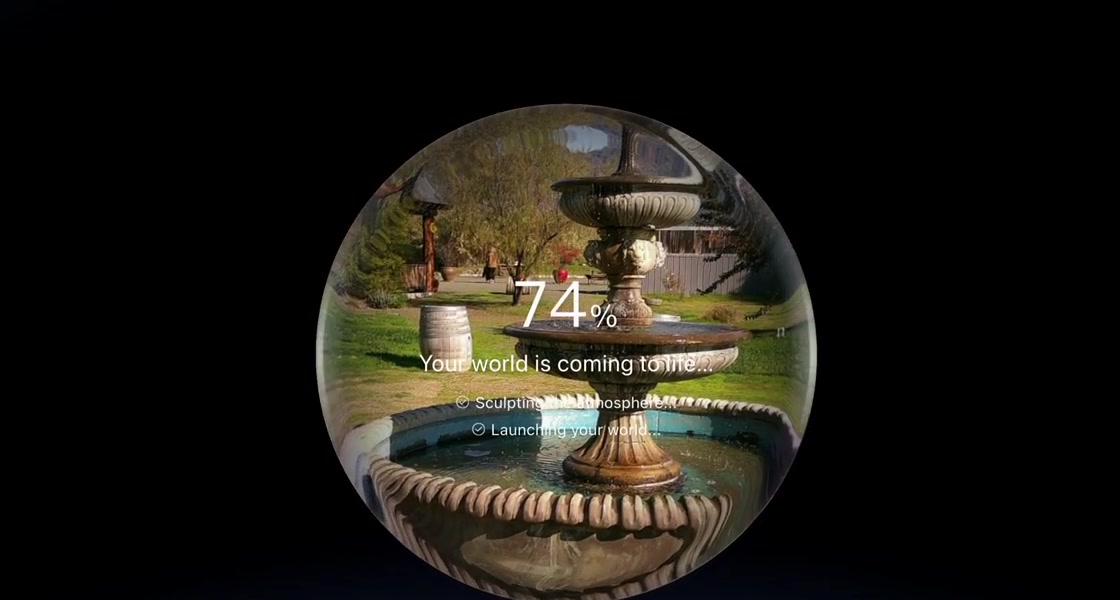}
    \includegraphics[width=0.32\linewidth]{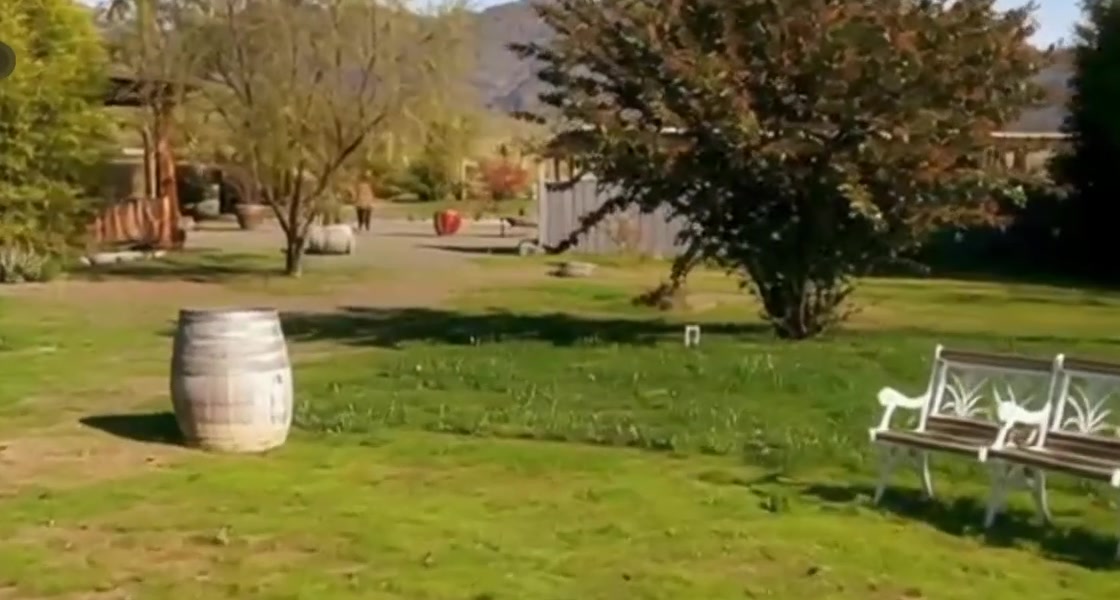}
    \\
    \vspace{0.15cm}
    \includegraphics[width=0.47\linewidth]{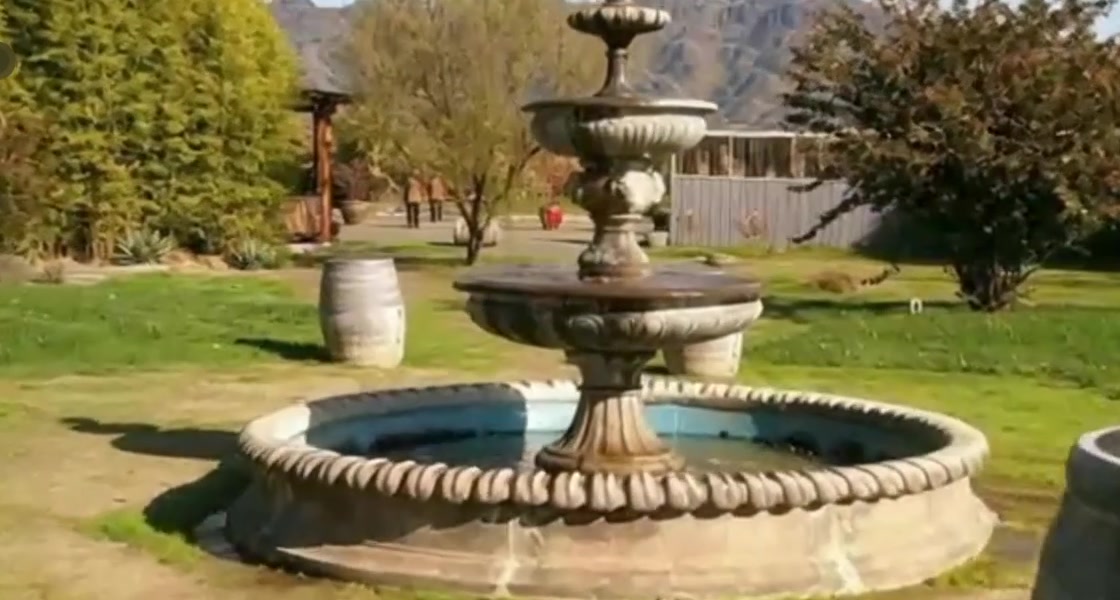}
    \includegraphics[width=0.47\linewidth]{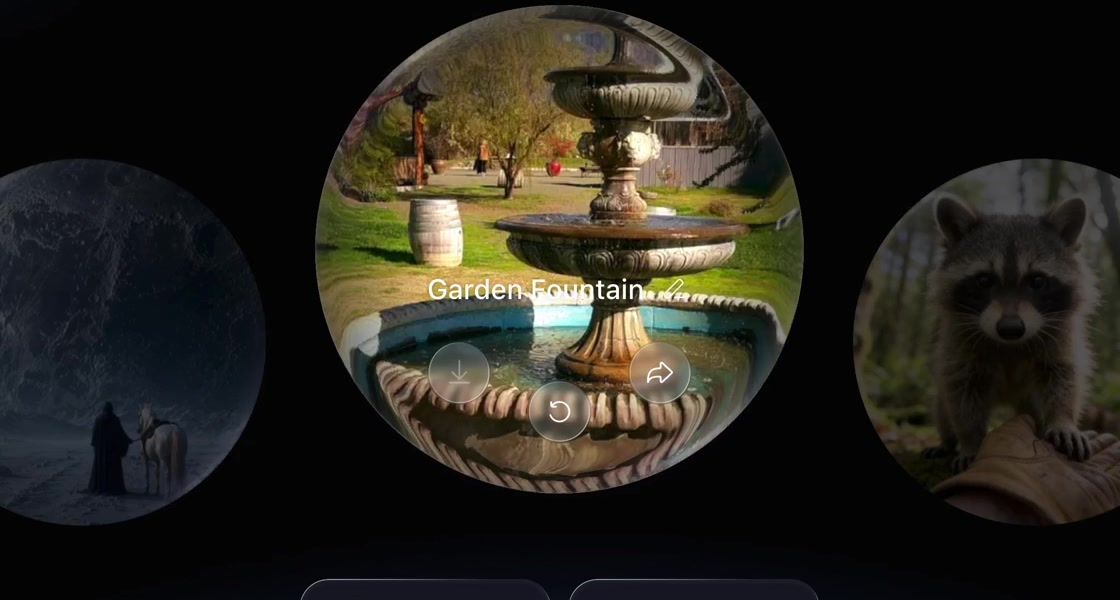}
    \caption{\textbf{Agent interface execution on a web-based world model.}
    (a) The agent interface enters the initial-world condition; (b) waits until the generated world becomes interactive; (c)--(d) dispatches controls and captures observations for the Agent Player's online decisions; and (e) terminates the episode while preserving the final video and execution record.}
    \label{fig:happyoyster_execution}
\end{figure}

\noindent\textbf{Example execution.}
Fig.~\ref{fig:happyoyster_execution} shows one HappyOyster case from caption entry to termination.
The screenshots are direct crops from the recorded browser session; browser chrome, system controls, and assistant overlays are excluded without altering the generated world content.

\section{Human Annotation and Validation}
\label{sec:supp_human_annotation}

\noindent\textbf{Benchmark annotation.}
Human annotators screen candidate initial worlds, verify Gemini-generated captions, assign the viewpoint, write long-horizon objectives and basic action sequences, and construct sample-specific VQA rubrics.
Annotators use observable evidence and remove questions whose answers cannot be determined from the generated rollout.

\noindent\textbf{Human-alignment study.}
Five participants independently evaluate the same 120 video pairs, comprising 30 comparisons for each of geometry consistency, interaction fidelity, out-of-sight evolution, and insight evolution.
This yields 600 valid pairwise judgments, with no missing responses.
For each comparison, participants view two videos generated for the same evaluation case, with model identities concealed using anonymous codes.
They select the left video, the right video, or \emph{Tie} when the two are difficult to distinguish.

Human preference assigns full credit to a win and half credit to a tie.
Overall human preference pools comparisons across the four equally sampled dimensions.
For visualization, we normalize the preference rates within each dimension so that the scores of the nine evaluated models sum to $100\%$.

\begin{table*}[!t]
    \caption{\textbf{Inter-rater agreement in the human-alignment study.}
    All results are computed from the same 120 pairwise items and 600 judgments used in Fig.~\ref{fig:human_alignment}.
    Unanimous Agreement requires all five raters to select the same outcome, whereas Majority Agreement requires at least three raters to agree.}
    \label{tab:human_agreement}
    \centering
    \renewcommand{\arraystretch}{1.15}
    \resizebox{0.88\linewidth}{!}{%
    \begin{tabular}{lccccc}
        \toprule
        Dimension
            & Items
            & Judgments
            & \makecell{Unanimous\\Agreement}
            & \makecell{Majority\\Agreement}
            & Fleiss' $\kappa$ \\
        \midrule
        Geometry Consistency & 30 & 150 & 33.3\% & 96.7\% & 0.461 \\
        Interaction Fidelity & 30 & 150 & 16.7\% & 96.7\% & 0.323 \\
        Out-of-sight Evolution & 30 & 150 & 23.3\% & 100.0\% & 0.419 \\
        Insight Evolution & 30 & 150 & 43.3\% & 90.0\% & 0.483 \\
        \midrule
        Overall & 120 & 600 & 29.2\% & 95.8\% & 0.434 \\
        \bottomrule
    \end{tabular}}
\end{table*}

At least three of the five raters agree on $95.8\%$ of all pairwise items.
Fleiss' $\kappa$ is $0.434$ overall and remains positive across all four dimensions, supporting the reliability of the aggregated human preference rankings.
Figure~\ref{fig:human_alignment} compares the normalized human preference scores with the corresponding rubric-based VQA scores.
The Spearman correlations are $\rho=0.933$ for Overall, $\rho=0.983$ for geometry consistency, $\rho=0.933$ for interaction fidelity, $\rho=0.812$ for out-of-sight evolution, and $\rho=0.745$ for insight evolution; all are statistically significant ($p<0.05$).

\end{document}